\documentclass[letterpaper, 10 pt, conference]{ieeeconf}
\IEEEoverridecommandlockouts
\usepackage{graphicx} 
\usepackage{amsmath} 
\usepackage{amssymb}  
\usepackage{multirow}
\usepackage{booktabs}
\usepackage[colorlinks,urlcolor=black,linkcolor=black,anchorcolor=black,citecolor=black]{hyperref}
\usepackage{subcaption}
\newtheorem{theorem}{Theorem}
\usepackage{algorithm}
\usepackage[noend]{algpseudocode}

\title{\LARGE \bf
AFR: An Efficient Buffering Algorithm for Cloud Robotic Systems
}

\author{Yu-Ping Wang$^{1, 2}$, Hao-Ning Wang$^{3}$, Zi-Xin Zou$^{1}$, Dinesh Manocha$^{4}$ \\
The source code and introduction video can be found at \url{https://github.com/Jrdevil-Wang/AFR}.
\thanks{$^{1}$Yu-Ping Wang and Zi-Xin Zou are with the Department of Computer Science and Technology, Tsinghua University, Beijing 100084, China. }%
\thanks{$^{2}$Yu-Ping Wang is also with the School of Computer Science and Technology, Beijing 100081, China. }%
\thanks{$^{3}$Hao-Ning Wang is with the Department of Mathematical Science, Tsinghua University, Beijing 100084, China. }%
\thanks{$^{4}$Dinesh Manocha is with the Department of Computer Science, University of Maryland, MD 20742, USA. }%
\thanks{Yu-Ping Wang is the corresponding author, e-mail: wyp@tsinghua.edu.cn.}%
\thanks{This work was supported by the National Natural Science Foundation of China under Grant No.~61872210.}%
}
\begin{document}

\maketitle
\thispagestyle{empty}
\pagestyle{empty}

\begin{abstract}

Communication between robots and the server is a major problem for cloud robotic systems.
In this paper, we address the problem caused by data loss during such communications, and propose an efficient buffering algorithm, called AFR, to solve the problem.
We model the problem into an optimization problem to maximize the received Quantity of Information (QoI).
Our AFR algorithm is formally proved to achieve near-optimal QoI, which has a lower bound that is a constant multiple of the unrealizable optimal QoI.
We implement our AFR algorithm in ROS without changing the API for the applications.
Our experiments on two cloud robot applications show that our AFR algorithm can efficiently and effectively reduce the impact of data loss.
For the remote mapping application, the RMSE caused by data loss can be reduced by about 20\%.
For the remote tracking application, the probability of tracking failure caused by data loss can be reduced from about 40\%-60\% to under 10\%.
Meanwhile, our AFR algorithm introduces time overhead of under 10 microseconds.

\end{abstract}

\IEEEpeerreviewmaketitle

\section{Introduction}

\begin{figure}[!t]
\centering
  \begin{subfigure}[t]{0.4\columnwidth}
    \centering
    \includegraphics[width=\textwidth]{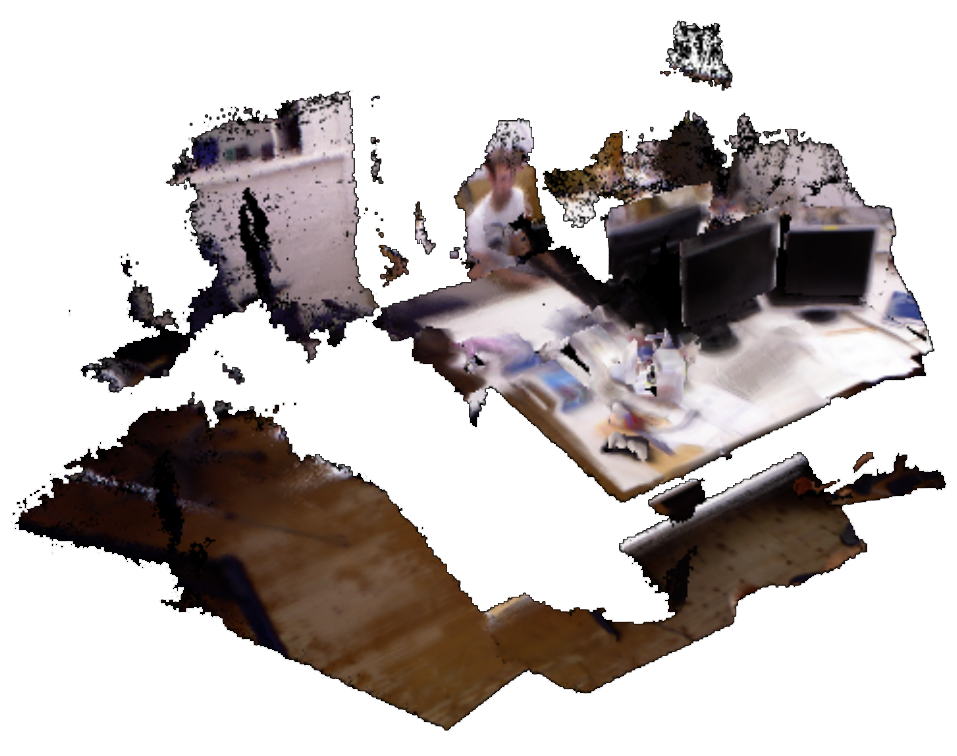}
    \caption{}
    \label{fig:shot_ref}
  \end{subfigure}
  \begin{subfigure}[t]{0.4\columnwidth}
    \centering
    \includegraphics[width=\textwidth]{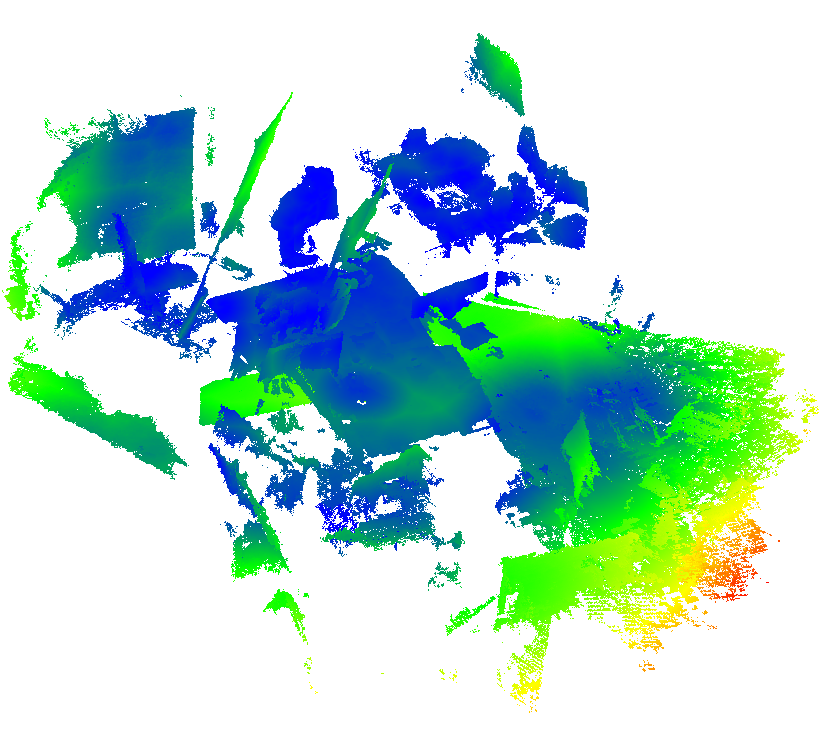}
    \caption{}
    \label{fig:shot_orig}
  \end{subfigure}
  \\
  \begin{subfigure}[t]{0.4\columnwidth}
    \centering
    \includegraphics[width=\textwidth]{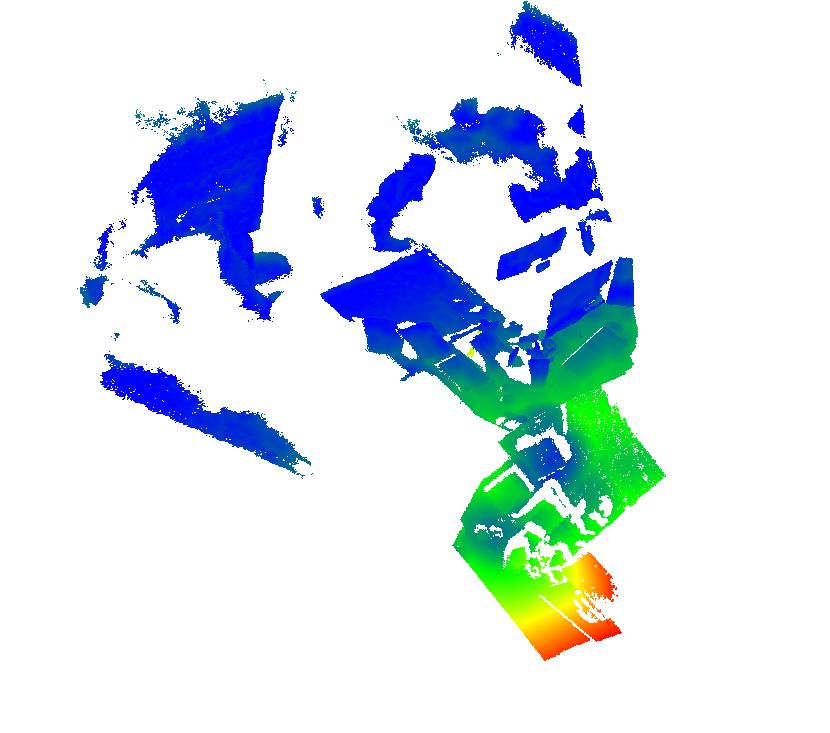}
    \caption{}
    \label{fig:shot_rand}
  \end{subfigure}
  \begin{subfigure}[t]{0.4\columnwidth}
    \centering
    \includegraphics[width=\textwidth]{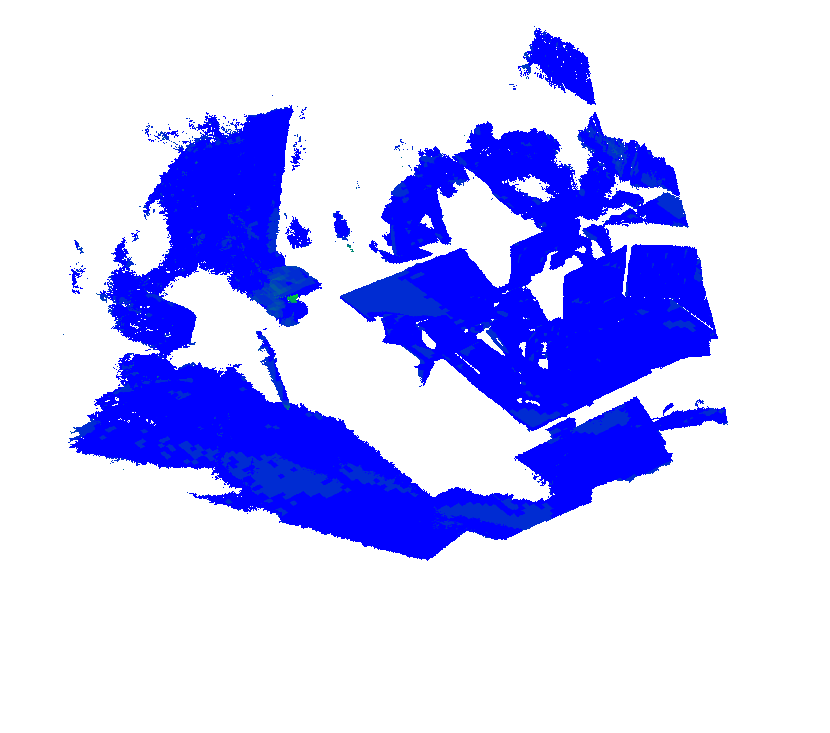}
    \caption{}
    \label{fig:shot_afr}
  \end{subfigure}
\caption{(a) The reference reconstructed 3D model. A person is sitting at a table with three screens on it. (b) Resulting 3D model when using the Drop Oldest algorithm. (c) Resulting 3D model when using the Random Drop algorithm. (d) Resulting 3D model when using our AFR algorithm. (b-d) are the results for the queue size $L$=15. They are generated by CloudCompare, where red points represent points with large error and blue points represent points with small error.}
\label{fig:shot}
\end{figure}

Modern robotic systems have become a substitute for humans when it is necessary to perform risky or exhausting tasks such as military operations, exploration, rescue operations, surveillance, or large-scale cleaning operations.
In such applications, robots need long-term autonomy.
Considering the cost and energy consumption, off-loading some tasks to a powerful server is a reasonable solution.
In addition, by gathering information from multiple robots, the server can make global decisions that a single robot cannot make.
Such an architecture is a cloud robotic system~\cite{DBLP:journals/rcim/DawarkaB22}.
Under such an architecture, communication between robots and the server is a major problem.

Due to the nature of autonomous robots, they usually communicate with the server over a wireless network (e.g., over WiFi or 5G).
However, wireless network connections are not always stable~\cite{DBLP:journals/icl/HooftPWHRBT16}.
The network could suffer temporary bandwidth reduction or even an interruption when the robot moves out of range of the wireless network, switches between network access points, or is obscured by obstacles (e.g., driving into a tunnel).
In this case, buffering some messages at the sender is a commonly used solution.
When network connections are temporarily broken, new messages are put into a buffer to wait for future transmission.
However, due to the limited memory of low-cost robots, the size of the buffer is limited.
Commonly used robotic middleware, such as Robot Operating System (ROS)~\cite{ros}, employ a policy called the Drop Oldest policy~\cite{DBLP:conf/colcom/ChenYZW16}.
When the buffer is full, this policy discards the oldest message to make room for the new message.
This solution is reasonable when newer messages are considered more valuable than older ones.

However, we find that this assumption does not hold for many robotic applications.
Taking the remote tracking application as an example, the robot sends large sensor data to the server, and the server replies with the robot's global location~\cite{DBLP:conf/icra/OpdenboschOGAS18}.
Discarding the oldest message would cause more differences between adjacent messages and may cause tracking failure.
Prior work~\cite{DBLP:conf/iros/WangZWDQM21} has considered the problem and provided a specialized solution for the tracking application by computing similarities between adjacent images.
Such similarity computations introduce time overhead.
We need a more efficient solution that can be generalized to other applications.
Taking the remote mapping application as an example, the robot sends large sensor data to the server, and the server integrates them into a dense 3D model map~\cite{DBLP:journals/tvcg/GolodetzCLPMT18}.
Discarding the oldest message would lose information that no other messages can provide, but comparing similarities between adjacent images is not reasonable for this application.

\textbf{Main Results:} In this paper, we design a novel network buffering algorithm, called AFR, to reduce the impact of data loss.
We model the \textit{Quantity of Information} (QoI) of the data received at the server, and our AFR algorithm will discard the message that has minimal impact on the received QoI.
By combining two frame rates and adaptively adjusting them, our AFR algorithm can near-optimally maximize the received QoI without estimating the network bandwidth or any feedback from the server, relying only on the status of the buffer.
The main contributions of this paper are summarized as follows.
\begin{itemize}
    \item We model the QoI of the data received at the server under the cloud robotic system architecture.
    Based on this model, we formalize the problem of a limited buffer to maximize the received QoI.
    \item We design a novel buffering algorithm, called AFR.
    We formally prove that our AFR algorithm is better than the Drop Oldest and the Random Drop algorithms, and that it is near-optimal with a bounded difference with the optimal, but unrealizable, Oracle algorithm.
    \item We implement our AFR algorithm based on ROS~\cite{ros}, which is the \textit{de facto} robotic middleware.
    Evaluations through practical applications confirm the superiority of our AFR algorithm compared to prior algorithms.
\end{itemize}

\section{Related Work}

\subsection{Communication Issues in Robotic Systems}

Communications between robots and the server have been a major issue in multiple-robot systems.
Saeedi et al. review multiple-robot systems~\cite{DBLP:journals/jfr/GTSL16}, concluding that communication is one of ten major problems.
Reducing the frame rate, rather than decreasing the resolution, is the most used solution to relieve limited network bandwidth.
Golodetz et al.~\cite{DBLP:journals/tvcg/GolodetzCLPMT18} aim to reconstruct dense 3D models with collaborative robots.
They explicitly point out the communication problem between robots and the control center.
Based on some experiments and evaluation, they compress the sensor images and sample the image frames to about 10 fps (frames per second). 
Dong et al.~\cite{DBLP:journals/tog/Dong0ZTXNC19} perform collaborative scanning for dense 3D reconstruction with multiple robots.
New sensor data are only collected when the robot moves more than a given distance or rotates more than a given angle.
The given distance and angle are decided by the speed of their robots.
In practice, manually deciding the resolution and frame rate requires considerable experimentation and parameter tweaking and cannot be easily adapted to dynamic environments with varying network bandwidths.


\subsection{Adaptive Video Analytics}

Video analytics is also an important application of robotic systems.
Chameleon~\cite{DBLP:conf/sigcomm/JiangABSS18} periodically searches for the optimal configuration for a video query, balancing accuracy and computation resources.
It ignores the varying network bandwidth but focuses on the differences between video content at different times.
AWStream~\cite{DBLP:conf/sigcomm/ZhangJRWL18} evaluates the performance of different configurations in an offline way to establish a relationship between bandwidth and accuracy.
Based on the profiles of configurations, AWStream chooses the configuration with the best accuracy from those satisfying the estimated bandwidth status.
JCAB~\cite{DBLP:conf/infocom/Wang0CQWX20} aims to jointly optimize configuration adaption while balancing accuracy, latency, and energy consumption.
Runespoor~\cite{DBLP:conf/infocom/Wang21} addresses the problem of tail accuracy.
They all rely on feedback from the server to guide the configurations of clients.
When a network interruption occurs, congestion signals from the server are blocked.
Even if the client decides to switch configurations without the guide of the server, it can only be triggered when queued items exceed a threshold.
At this time, switching configurations for future messages cannot avoid the necessity of dropping some newer messages.
Our AFR algorithm would be a complement to these frameworks.

\subsection{Quality of Service}

Quality of Service (QoS) policies are designed to meet the needs of different scenarios, such as real-time requirements.
However, existing QoS policies are not suitable for our problem of cloud robotic systems.
ROS2 provides QoS policies by integrating with Data Distribution System (DDS)~\cite{DBLP:conf/icdcsw/Pardo-Castellote03}.
According to the specifications~\cite{ros2_qos2} and evaluations~\cite{ros2_eval}, 
the \textit{LatencyBudget} configuration is the most related policy because it implies the maximal frequency.
However, deciding the maximal frequency needs more experiments~\cite{DBLP:journals/tvcg/GolodetzCLPMT18} and is vulnerable to temporary network interruption.

\begin{figure*}[!t]
\centering
\includegraphics[width=0.65\textwidth]{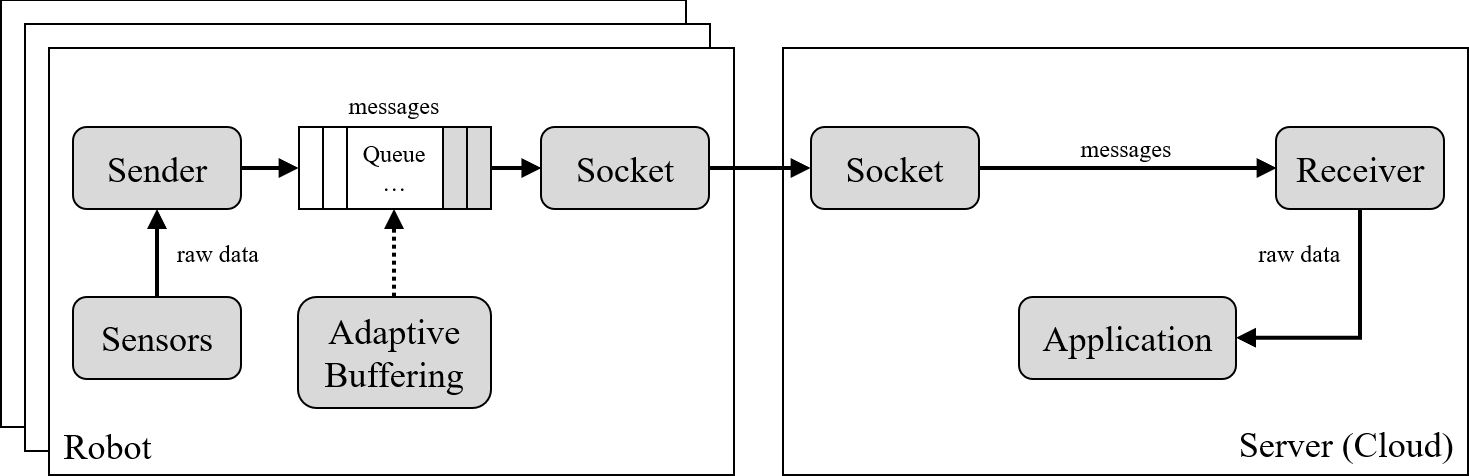}
\caption{The architecture of cloud robotic systems.}
\label{fig:arch}
\end{figure*}

\subsection{Buffering Algorithms}

Buffering algorithms have been researched for the last decade.
Practice networks are modeled as Delay Tolerant Networks (DTN)~\cite{DBLP:journals/comsur/CaoS13} or Opportunistic Networks~\cite{DBLP:conf/icccn/SatiPG16}.
A sequence of simple drop schemes such as Drop Oldest, Drop Youngest, Drop Front, and Drop Last~\cite{DBLP:conf/colcom/ChenYZW16} has been proposed when the buffer is full.
Enhanced policies introduce the concept of \textit{profit} and assume that each message has a different profit.
The goal is to maximize the total profit of the remaining messages~\cite{DBLP:journals/sigact/Goldwasser10}.
According to the application scenarios, different types of messages may have different profits~\cite{DBLP:conf/csa2/KimW17,DBLP:journals/pomacs/YangWH17,DBLP:conf/apcc/KimuraM17}.

However, we find that in many robotic applications, the profit of a message is not decided by itself.
To address this aspect, we have proposed ORBBuf~\cite{DBLP:conf/iros/WangZWDQM21}.
By defining similarity between messages, ORBBuf discards the message that has minimal impact on the continuity between images, which is crucial to the success of SLAM algorithms.
However, ORBBuf is specially designed for SLAM algorithms, and introduces time overhead to compute the similarity.
In this paper, we propose a novel buffering algorithm for more general applications by modeling the QoI.
To the best of our knowledge, ours is the first work that models the QoI based on robot movement.

\section{Modelling Cloud Robotic Systems}

Figure~\ref{fig:arch} shows the general workflow of cloud robotic systems.
At the robot side, the \textit{Sender} module gathers raw data (e.g., color images, depth images, infrared images, etc.) from \textit{Sensors}, and compresses the data gathered at one time into a message.
The message is then enqueued to a \textit{Queue} and waits for transmission.
The \textit{Socket} repeatedly dequeues a message from the \textit{Queue} and transmits it through the wireless network.
These messages are usually large in size (e.g., a single compressed $640 \times 480$ depth image takes about 200KB of memory~\cite{DBLP:conf/tabletop/Wilson17}) and must be split into hundreds of packages (e.g., 1500 bytes per package) during transmission.
Therefore, unreliable network protocols such as UDP are not preferred in this case, and we use TCP as the underlying transmission protocol.
At the server side, the \textit{Receiver} gets a transmitted message, decompresses it into raw data, and provides the raw data to robotic applications.

When a network interruption occurs, the \textit{Socket} dedicates itself to the transmission of a single message, and the \textit{Queue} grows, until it reaches its limit.
Then, \textit{Buffering} algorithm decides whether we should discard the newest message, or which message in the queue should be discarded to make room for the newest message.
In either case, the process of discarding a message results in data loss, but we can optimize the buffering algorithm so that the overall \textit{Quantity of Information} (QoI) of all received messages is maximized.
Obviously, the key to defining our problem is to model the received QoI.

\subsection{Modeling the Quantity of Information}

C. E. Shannon defined the quantity of information with entropy, i.e., the uncertainty of random events.
We follow the notation used in~\cite{DBLP:journals/bstj/Shannon48} to model our problem.
If the \textit{Receiver} receives a sequence of $n$ messages $M=\{M_0, M_1, ..., M_{n-1}\}$, the overall QoI $H(M)$ is:
\begin{equation}\label{equ:qoi}
    H(M) = H(M_0) + \sum_{i=1}^{n-1} H(M_i | M_0, ..., M_{i-1})
\end{equation}
where $H(x | y)$ is the entropy of $x$ with known $y$.
Indeed, there are usually overlaps between the information provided by messages, especially adjacent messages.

\begin{figure}[!t]
\centering
  \begin{subfigure}[t]{0.35\columnwidth}
    \centering
    \includegraphics[width=\textwidth]{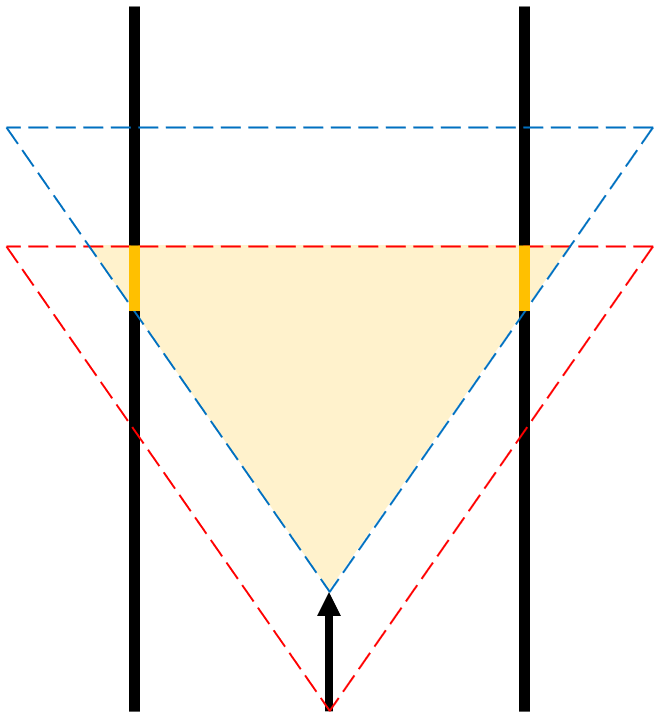}
    \caption{}
    \label{fig:movement1}
  \end{subfigure}
  \begin{subfigure}[t]{0.5\columnwidth}
    \centering
    \includegraphics[width=\textwidth]{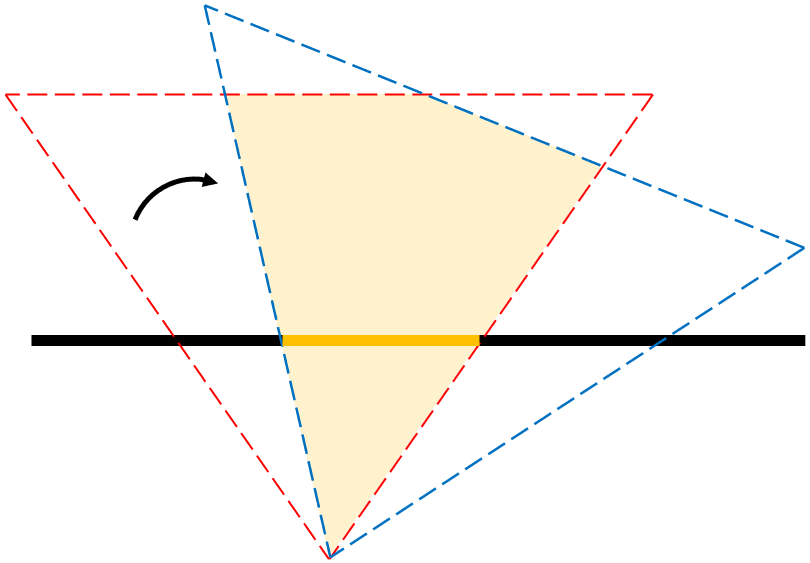}
    \caption{}
    \label{fig:movement2}
  \end{subfigure}\\
  \begin{subfigure}[t]{0.48\columnwidth}
    \centering
    \includegraphics[width=\textwidth]{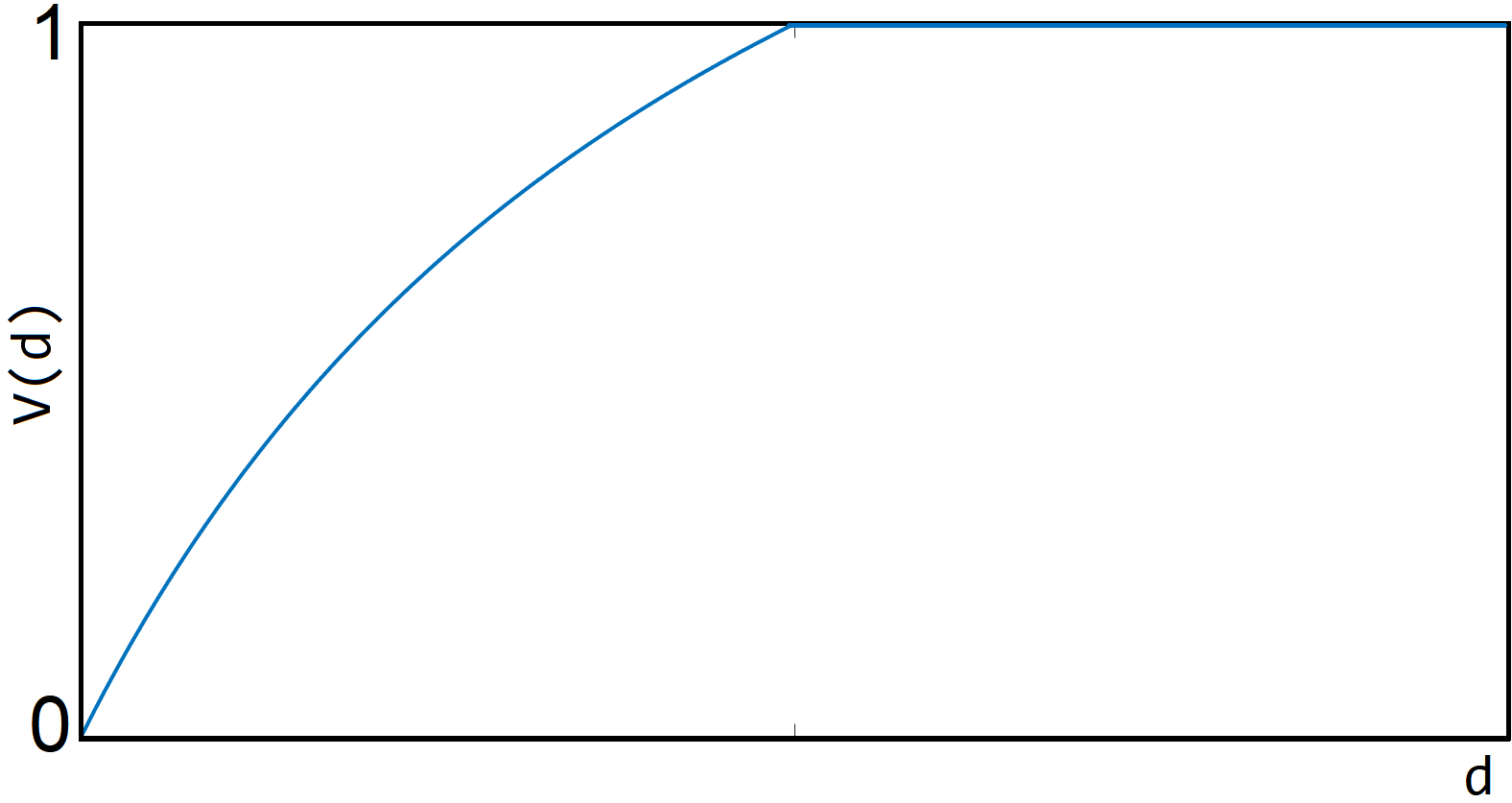}
    \caption{}
    \label{fig:forward}
  \end{subfigure}
  \begin{subfigure}[t]{0.48\columnwidth}
    \centering
    \includegraphics[width=\textwidth]{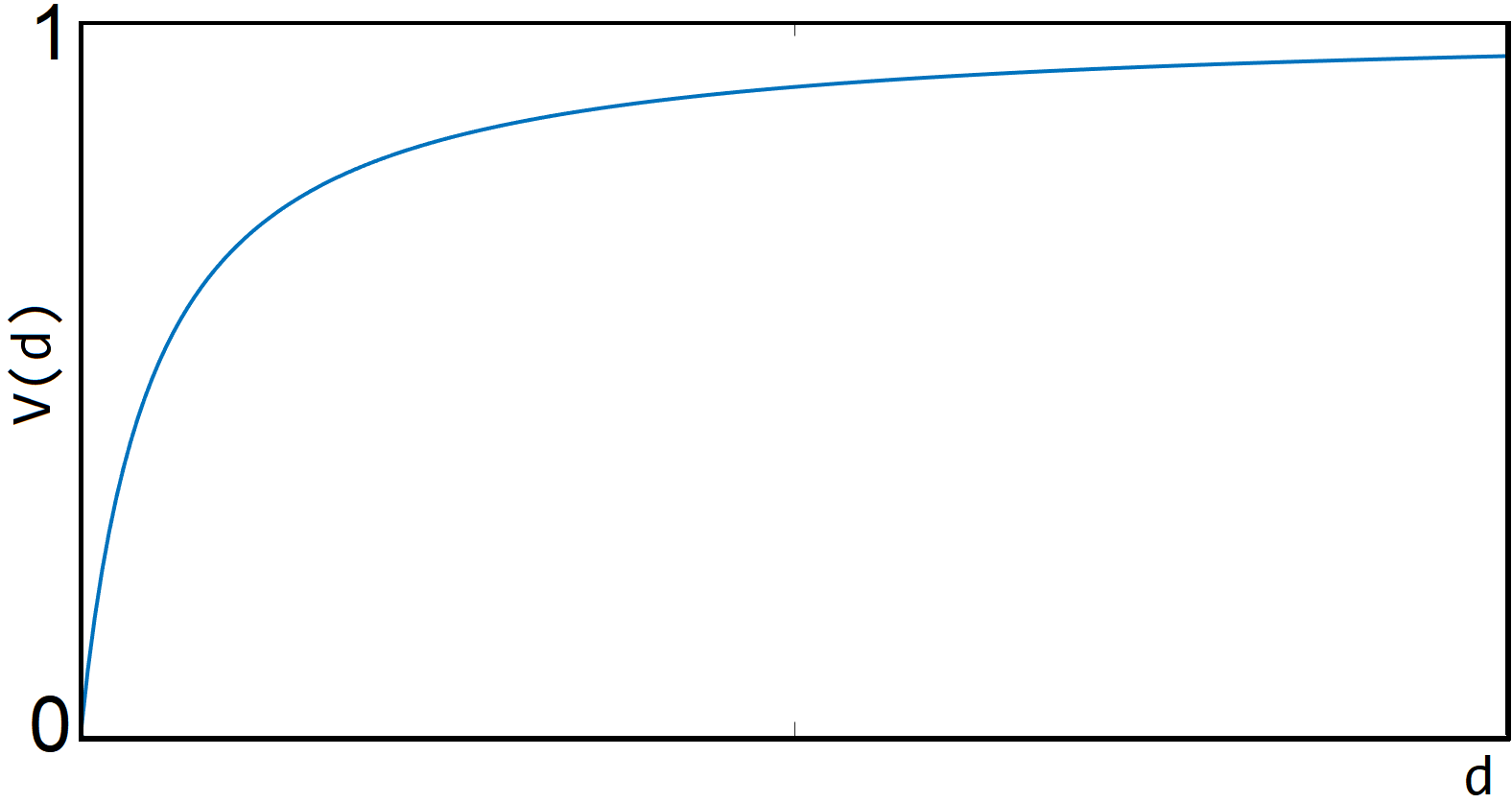}
    \caption{}
    \label{fig:turn}
  \end{subfigure}
\caption{Two examples of the overlap between depth images while the robot moves. The red triangle represents the field of view (FoV) at the former position, and the blue triangle represents the FoV at the latter position. The black lines represent real-world walls, and the yellow parts are the overlaps between two depth images. (a) The robot is moving forward along a corridor. (b) The robot is turning around in place. (c) The corresponding $V$ function of (a). (d) The corresponding $V$ function of (b). Note that (c) and (d) satisfy our assumptions in Equations~\ref{equ:q0}-\ref{equ:qconcave}.}
\label{fig:movement}
\end{figure}

For better understanding, we use the application of building a map with a depth camera to illustrate the problem, and other applications have similar forms.
Figure~\ref{fig:movement} shows two examples of the overlap between depth images while the robot moves.
The depth camera can capture the distance between the camera and objects within its field of view (FoV).
The FoV of a depth camera is a 3D region in the shape of a rectangular pyramid (represented with a triangle in Figure~\ref{fig:movement}) containing a set of rays from the camera.
Some of the rays hit an object, and the camera measures the axial distance between the object and itself.
The other rays that do not hit any objects indicate that there are no objects along these rays within the range limit.
In Figure~\ref{fig:movement1}, the robot is moving forward along a corridor.
At the former position (the vertex at the bottom of the red triangle), a part of the walls is within the FoV; at the latter position (the vertex at the bottom of the blue triangle), another part of the walls is within the FoV.
Their intersection is redundant overlap, which is highlighted in yellow.
The depth image provided by the camera at the latter position provides less new information, because we have already been certain of the highlighted part of walls based on the information at the former position.
Similarly, in Figure~\ref{fig:movement2}, the robot is turning around in place, and the intersection is highlighted.

From the above examples, we can see that the actual $H(M_i|M_0,...,M_{i-1})$ is complicated, and is related to the positions of the camera and objects.
However, by summarizing some common cases, we can simplify it with some assumptions.

Assumption 1: \textit{The information provided by every $M_i$ is the same}.
This is reasonable since the resolutions of images remain the same. 
Therefore, without loss of generality, we assume that $H(M_i)=1$.

Assumption 2: \textit{The information provided by $M_i$ but not by $M_{i-1}$ is not provided by any $M_j$ with $j < i$}.
This assumption may not be true if the robot goes back.
However, it is more likely to be true in a short period of time, such as during a network interruption.
Thus, we can simplify $H(M_i|M_0,...,M_{i-1})$ to $H(M_i|M_{i-1})$.

Assumption 3: \textit{$H(M_i|M_{i-1})$ is larger when the time difference between them is larger}. This is reasonable since larger time differences indicate less probability and quantity of their overlap. If the time stamp of $M_i$ is $T_i$, we further simplify $H(M_i|M_{i-1})$ to a function $V$ of their time difference, i.e., $V(T_i - T_{i-1})$.

Overall, we can simplify Equation~\ref{equ:qoi} to:
\begin{equation}\label{equ:qois}
    H(M)=1+\sum_{i=1}^{n-1}V(T_i-T_{i-1})
\end{equation}

We further model a variety of simple movement patterns of the robot (Figure~\ref{fig:movement} shows two of them).
Although the resulting functions $V$s vary, they all follow the following properties.
\begin{equation}\label{equ:q0}
    V(0) = 0
\end{equation}
\begin{equation}\label{equ:qlimit}
    \forall d \geq 0, V(d) \leq 1
\end{equation}
\begin{equation}\label{equ:qmono}
    V\text{ is monotonic non-decreasing}.
\end{equation}
\begin{equation}\label{equ:qconcave}
    V\text{ is non-convex}.
\end{equation}

Equation~\ref{equ:q0} indicates that the same message provides no new information.
Equation~\ref{equ:qlimit} is reasonable under assumption 1.
Equation~\ref{equ:qmono} is reasonable under assumption 3.
Equation~\ref{equ:qconcave} seems a little strong, but we consider it reasonable under Equations~\ref{equ:q0}-\ref{equ:qmono}, and we find that it is true for a variety of simple robot motions (including the two examples shown in Figure~\ref{fig:movement}).

\subsection{Problem Definition}

Suppose there are $T$ messages generated during the network interruption.
Without loss of generality, we assume their time stamps are $1, 2, ..., T$.
The message $0$ has been transmitted before the network interruption, or is being transmitted when the network interruption occurs.
In the latter case, the TCP protocol promises that the message $0$ will be successfully transmitted after the network resumes.
Messages from $T+1$ and after will all be transmitted once the network resumes.
When the upper limit of the queue is $L$, our problem is to choose at most $L$ messages from the $T$ messages and maximize the overall QoI.
Based on the model of QoI, we can formulate our problem as follows:
\begin{equation}\label{equ:problem}
\begin{split}
                 &arg\max_{M} H(\hat{M}) {}\\
    \text{ s.t. }&\hat{M} = \{0\} \uplus M \uplus \{T+1\}, {}\\
                 &|M| \leq L,~M \subset \{1, 2, ..., T\}
\end{split}
\end{equation}
where $\uplus$ indicates concatenating two sequences, and $|M|$ indicates the length of sequence $M$.

Obviously, if $T \leq L$, there is enough space in the queue to keep all $T$ messages.
This case indicates that, in order to maximize the overall QoI, we should \textbf{NOT} drop any messages before the queue fills up.
In the rest of the paper, we focus on the case of $T > L$.
To avoid misunderstanding, we use ``\textit{result}" to refer to the resulting maximal QoI, and ``\textit{solution}" to refer to the resulting message sequence $M$.

\subsection{Oracle Algorithm}\label{sec:oracle}

To solve the problem stated in Equation~\ref{equ:problem}, we design a powerful algorithm called $Oracle$.
According to the Oracle algorithm, we chose $L$ messages whose time stamps are $\{\delta, 2\delta, ..., L\delta\}$ where $\delta=\frac{T+1}{L+1}$.
We can prove that this algorithm is the \textit{optimal} solution, based on the property of function $V$ stated in Equations~\ref{equ:q0}-\ref{equ:qconcave}.
\footnote{See Appendix~\ref{sec:proof} for all the proofs.}

\begin{theorem}\label{thm:best}
No solution is better than that of $Oracle$.
\end{theorem}

However, this algorithm is unrealizable for two reasons.
First, $\delta$ might not be an integer, which violates the constraint in Equation~\ref{equ:problem}.
Second, even if we round them to their nearest integers and get an approximate optimal solution, we do not know how long the network interruption will last, which means we do not know $T$ in advance.
For example, when $L=4$ and $T=9$, the Oracle algorithm's output is $\{2, 4, 6, 8\}$; when $T=14$, the Oracle algorithm's output is $\{3, 6, 9, 12\}$.
However, the Oracle algorithm dropped message $3$ at time $T=9$, and cannot retrieve it in practice.
There is an extra constraint for a practical algorithms that dropped messages cannot be retrieved.
We can formulate this constraint by treating the solution $M$ as a function of $T$, and:
\begin{equation}
    M(T) \subset M(T-1) \uplus \{T\}
\end{equation}

Although the Oracle algorithm is unrealizable, it can help us realize the upper bound of practical algorithms.
In addition, it also shows that we should generate a message sequence with a uniform time distance between each adjacent pair, i.e., a solution with a uniform frame rate.
This cannot be achieved because the frame rate changes as $T$ increases, and we do not know $T$ in advance.
Our idea is to use two frame rates to approximate the desired frame rate.

\section{Our AFR Buffering}\label{sec:algo}

In this section, we propose our buffering algorithm, \textit{AFR} (Adaptive Frame Rate), analyze its characteristics, and introduce our implementation.

\subsection{Algorithm Design}

The main idea of our AFR algorithm is to adjust the frame rate, not only for messages to be generated, but also for messages already in the queue.
At any time $T$, there are two frame rates adopted.
The queue is logically split into two parts.
After a \textit{dropping position} $p$, we keep 1 message for every $r=2^n$ messages; and before $p$, we keep 1 message for every $2^{n+1}$ messages.
We call this ratio $r$ for \textit{dropping rate}.
Note that it is easy to change the dropping rate from $r$ to $2r$, by further dropping 1 message for every 2 messages.

\begin{figure*}[!t]
\centering
\includegraphics[width=0.75\textwidth]{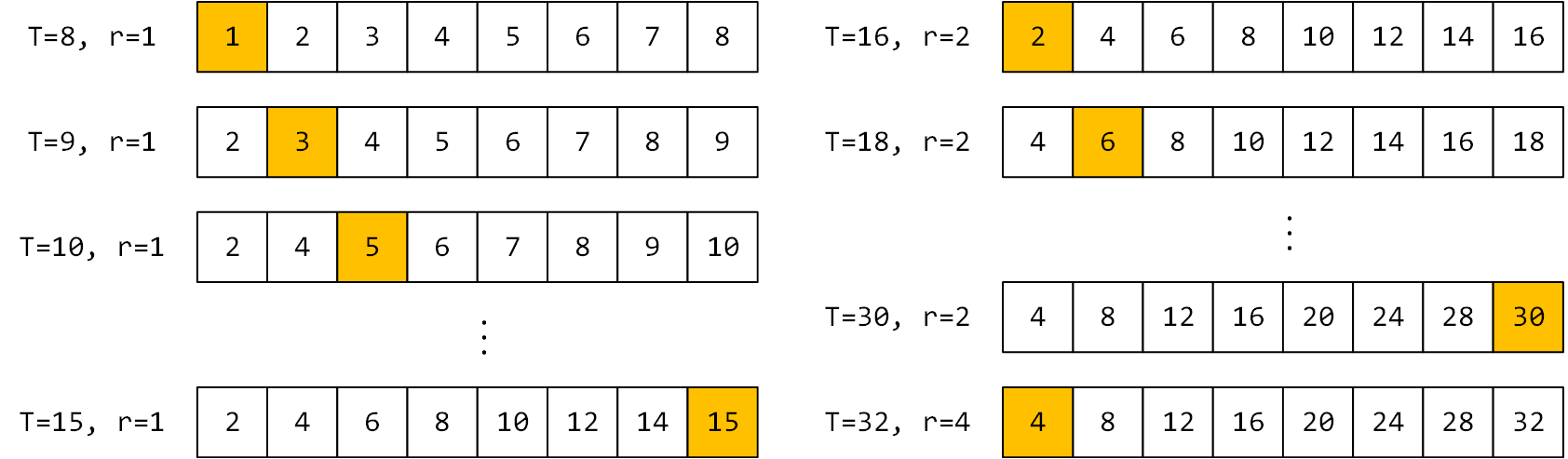}
\caption{An example of our AFR algorithm ($L = 8$). When a new message arrives and the queue is full, the message at the dropping position (highlighted) is dropped, and the dropping position moves to the next position. When the dropping position is at the end of the queue (e.g., when $T = 16, 32$), the frame rate $r$ is increased and the dropping position is moved back to the beginning.}
\label{fig:afr_example}
\end{figure*}

For better understanding, we will explain our AFR algorithm with an example, as shown in Figure~\ref{fig:afr_example}.
Suppose $L=8$ and we start by keeping every message ($r=1$).
At time $T=8$, the queue becomes full.
At time $T=9$, we must decide whether to drop message $9$ or make room for it.
Our AFR algorithm decides to drop message $1$, and label message $3$ as the next dropping position (the highlighted position).
At time $T=10$, AFR decides to drop message $3$, and label message $5$ as the next dropping position.
The process continues until time $T=16$, when AFR decides to drop message $15$, raise the dropping rate $r$ to 2, and reset the dropping position to message $2$.
From then on, AFR drops 1 message for every $r=2$ new messages.
Thus, at any time $T$, the dropping rate for messages before $p$ is $2r$, and the dropping rate for messages after $p$ and new messages is $r$.

During network interruption, the dropping rate $r$ keeps increasing.
When the network resumes, we should reduce the dropping rate.
This is achieved by checking whether the occupation ratio $q$ of the queue is lower than a threshold $R$.
In practice, we empirically set $R=\frac{2}{3}$.

Overall, our AFR algorithm is described as Algorithm~\ref{alg:afr}.
In lines 2-4, we detect and reduce the dropping rate; in lines 5-8, we drop the new message based on the dropping rate; in lines 10-14, we make room for the new message.
As we can see, the complexity of our AFR algorithm is $O(1)$ since there are no loops.
We can implement the queue with a linked list, or a vector pointing to messages to avoid copying messages.
In practice, our AFR algorithm is efficient enough and can finish under 10 microseconds, even on embedded devices.

\begin{algorithm}[h]
\caption{AFR Algorithm}\label{alg:afr}
\begin{algorithmic}[1]
    \Procedure{Enqueue}{message $m$} \Comment{$r$: dropping rate}
    \If {$r > 1$ \textbf{and} $q < R$} \Comment{$q$: occupation ratio}
        \State $r \gets r / 2$ \Comment{$R$: threshold}
        \State $c \gets c \mod r$
    \EndIf
    \State $c \gets c+1$ \Comment{$c$: message counter}
    \If {$c <> r$} \Comment{Drop frame based on $r$}
        \State \textit{drop} $m$, \Return
    \EndIf
    \State $c \gets 0$
    \If {\textit{queue is full}}
        \State \textit{drop the message at} $p$ \Comment{$p$: dropping position}
        \State $p \gets p+1$
        \If {$p = L$} \Comment{$L$: size limit of the queue}
            \State $r \gets r * 2$ \Comment{Half the frame rate}
            \State $p \gets 0$
        \EndIf
    \EndIf
    \State \textit{keep} $m$, \Return
    \EndProcedure
\end{algorithmic}
\end{algorithm}

\subsection{Analysis and Comparison}

In this subsection, we will analyze the characteristics of our AFR algorithm by comparing it with other algorithms.

\subsubsection{Oracle Algorithm}

As we introduced in Section~\ref{sec:oracle}, the Oracle algorithm is an optimal, but unrealizable, algorithm.

Since we have proved in Theorem~\ref{thm:best} that no solution is better than that of Oracle, our AFR algorithm cannot be better.
However, we can prove that in some cases, the solution of our AFR algorithm is the same as Oracle's.
In addition, there is a lower bound of the result of our AFR algorithm which is a constant multiple of the optimal result.

\begin{theorem}
$2(\sqrt{2}-1)\frac{L}{L+1} < \frac{H_{AFR}(T, L)}{H_{Oracle}(T, L)}\leq 1$,
\end{theorem}
where $H_{Oracle}(T, L)$ indicates the received QoI when using the Oracle algorithm at the time $T$ with buffer length $L$, and $H_{AFR}$ indicates that of our AFR algorithm.

\subsubsection{Drop Oldest Algorithm}

The Drop Oldest algorithm drops the oldest message in the queue when the queue is full.
It is very easy to implement and is probably the most used algorithm in practice.
For our problem, it generates a solution $M(T)=\{T-L+1, T-L+2, ..., T\}$.
We formally prove that our AFR algorithm is better than the Drop Oldest algorithm.

\begin{theorem}
$H_{AFR}(T, L)\geq H_{OLD}(T, L)$,
\end{theorem}

\subsubsection{Random Drop Algorithm}

The Random Drop algorithm drops a random message in the queue when the queue is full.
Each message in the queue has an equal probability of being dropped.
As a result, it is possible for this algorithm to generate every possible solution, including the best and the worst.
Intuitively, it seems that the \textit{average} solution fits our idea of uniform time distance.
However, this is a misguided conclusion because the Random Drop algorithm is not randomly dropping $T-L$ messages from $T$ messages but instead randomly dropping \textbf{$1$} message from $L+1$ messages for $T-L$ times.
The probability that the message $i$ is still in the queue after all dropping operations is:

\begin{equation}
    P\{i \in M(T)\}=(\frac{L}{L+1})^{min(T-i+1,T-L)}
\end{equation}

This indicates that older messages are much less likely to be kept in the queue than newer ones.
Thus, even the \textit{average} solution does not have a uniform time distance.
Unfortunately, without further assumptions about the function $V$, we cannot prove that our AFR algorithm is better than the Random Drop algorithm.
Instead, we run some simulations to support our claim.

\begin{figure}[!t]
\centering
\includegraphics[width=0.8\columnwidth]{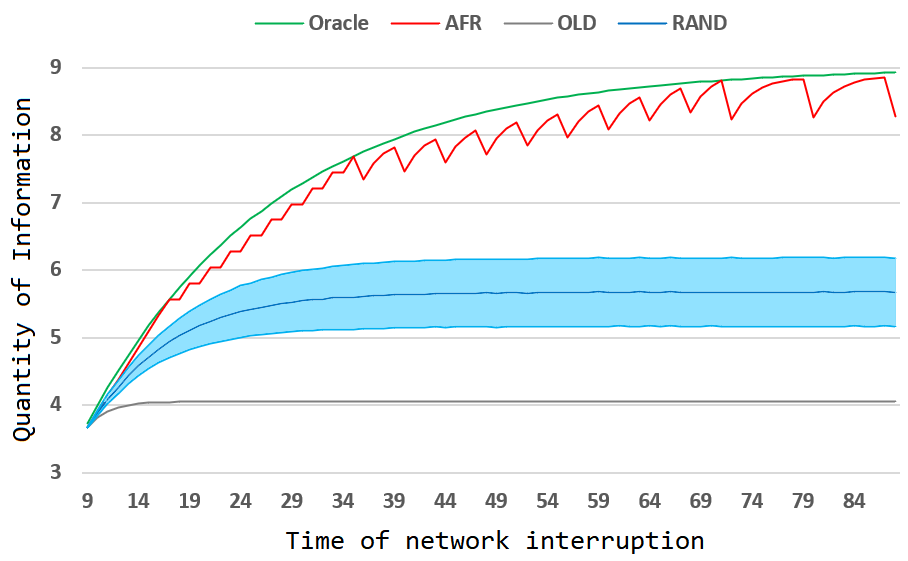}
\caption{Simulation results of different algorithms ($L = 8$).}
\label{fig:simulation}
\end{figure}

Figure~\ref{fig:simulation} shows the simulation results of different algorithms for $L=8$.
The horizontal axis is $T$ and the vertical axis is the resulting QoI.
In this simulation, we set the function $V(d)=1-0.618^d$.
This is the case when the robot is moving backward away from a wall.
Note that with $T$ increasing, the result of the Oracle algorithm (the green curve) increases, but never reaches $9$.
This is because with $L=8$, the QoI could reach $9$ if and only if there are no overlaps between each pair of adjacent messages (including $M_L$ and $T+1$).
The result of our AFR algorithm (the red curve) is very close to that of the Oracle algorithm.
In some special cases ($T=18, 36, 72...$), the result of our AFR algorithm is the same as that of the Oracle algorithm.
The result of the Drop Oldest algorithm (the grey curve) is very bad because it always keeps the $8$ newest messages, and there are a lot of overlaps among them.
The result of the Random Drop algorithm is random.
We run the simulation 10,000 times, and record the statistics.
The dark blue curve shows the expectation of all results, and the light blue region shows plus or minus 1 standard deviation (about 68\% results are within this region).
The \textit{average} result of the Random Drop algorithm is better than that of the Drop Oldest but worse than that of Oracle and our AFR.
This result confirms our analysis of the Random Drop algorithm.

\subsection{Implementation}

We implement our AFR algorithm based on ROS~\cite{ros}, which is the \textit{de facto} robotic middleware.
We modify the code of the class \textit{TransportSubscriberLink} which is located at the ROS Communication (\textit{ros\_comm}) project~\cite{ros_comm}.
The original \textit{ros\_comm} implements the Drop Oldest algorithm.
For comparison, we also modify it to implement the Random Drop algorithm and the ORBBuf algorithm~\cite{DBLP:conf/iros/WangZWDQM21}.
Each modified version is linked into a dynamic-link library \textit{libroscpp.so}.
Since we implement different algorithms at the middleware layer, they are transparent to the developers and no user code needs to be modified.
The only operation is to replace \textit{libroscpp.so} with the desired version.
It is also easy to provide an interface for developers to choose the desired algorithm, which is left for future work.

\section{Evaluation}

In this section, we show the advantage of our AFR algorithm with mapping and tracking applications.
\begin{itemize}
    \item Section~\ref{sec:eval-gt} shows that, by using our AFR algorithm, the error of the resulting map can be reduced by about 10\%-20\%.
    \item Section~\ref{sec:eval-ngt} shows that, by using our AFR algorithm, the probability of tracking failure is reduced from about 40\%-60\% to under 10\%.
    \item Section~\ref{sec:eval-time} shows that our AFR algorithm is efficient. The time overhead is under 10 microseconds, while the ORBBuf algorithm introduces about 10 milliseconds of overhead.
\end{itemize}

\subsection{Experiment Setup}

In the following experiment, each data sequence is replayed on a laptop that acts as the robot.
The server is equipped with an Intel Core i7-8750H @2.20GHz 12x CPU, 16GB memory, and an NVIDIA GeForce GTX 1080 GPU.
All software modules are connected with ROS middleware.
The ROS version is Lunar on Ubuntu 16.04.

We set up the experiments based on InfiniTAM~\cite{InfiniTAM_ISMAR_2015}, which is one of the most integrated state-of-the-art 3D reconstruction frameworks.
Its interface allows us to change the underlying mapping and tracking algorithms.

We use the TUM RGB-D dataset~\cite{sturm12iros} as input.
The dataset contains color and depth images captured with a Microsoft Kinect sensor which is a typical RGB-D sensor used in robotic systems.
The data was recorded at full frame rate (30 Hz) and sensor resolution ($640\times480$).
The dataset also provides the ground-truth trajectory of the sensor, which is obtained from a high-accuracy motion-capture system with external tracking cameras.

\begin{figure}[!t]
\centering
\includegraphics[width=0.8\columnwidth]{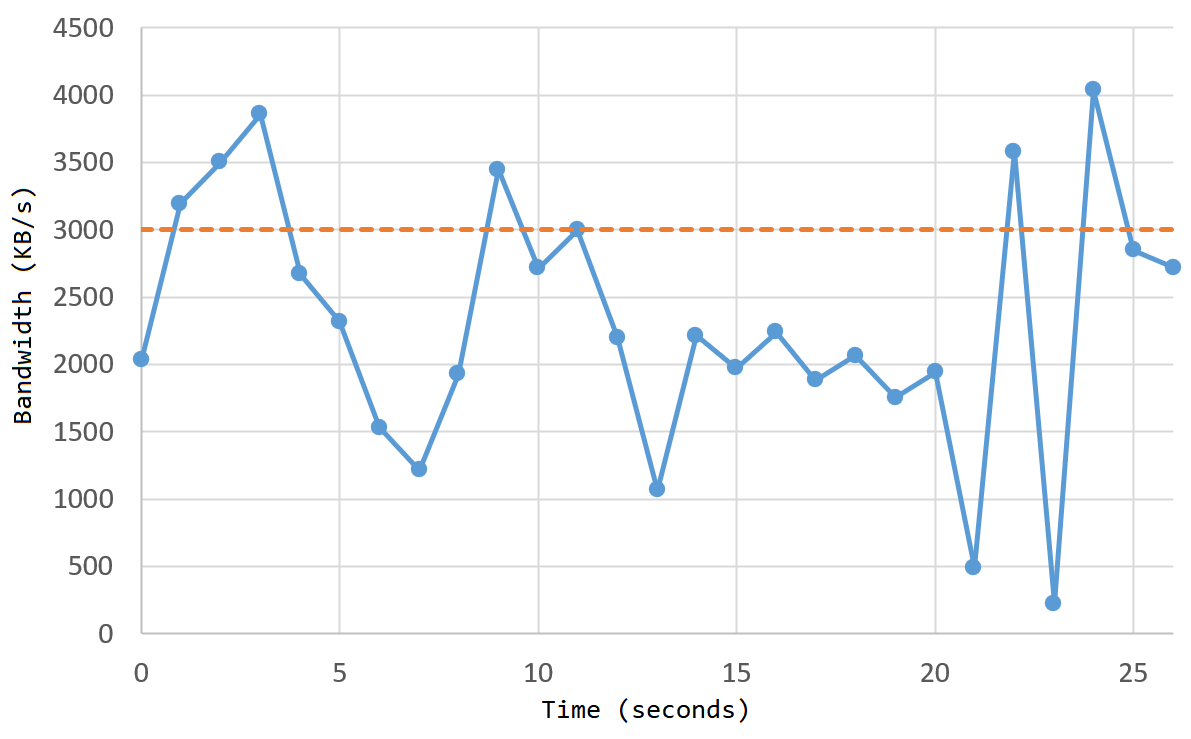}
\caption{The network bandwidth trace.}
\label{fig:bw}
\end{figure}

To reproduce the result multiple times, the robot and the server are connected directly with a 1Gbps Ethernet cable, and we use the Linux \textit{tc} utility to modify the outgoing bandwidth of the robot to emulate the cloud robotic environment.
We employ the Belgium 4G/LTE dataset~\cite{DBLP:journals/icl/HooftPWHRBT16}.
We choose a very challenging trace labeled \textit{car\_0002}, shown in Figure~\ref{fig:bw}.
The horizontal axis is time in seconds, and the vertical axis is the bandwidth in KB/s.
The orange dotted line indicates the approximate required bandwidth to transmit messages.
Generally, the bandwidth limit of a 4G network is about 15MB/s, which is enough to transmit all messages.
However, in this challenging trace, the network bandwidth varies second-wise, and the overall bandwidth is not enough to transmit all messages.
Specifically, at the 21st and 23rd second, the bandwidth drops below 500KB/s.

All test cases are repeated 10 times, and we organize and report the statistical result of each test case.


\subsection{Remote Mapping Application}
\label{sec:eval-gt}

In this experiment, we use the ground-truth trajectory to guide the dense mapping algorithm~\cite{DBLP:journals/tog/NiessnerZIS13}.
Each message contains a color image (compressed with low-quality JPEG), a corresponding depth image (compressed with lossless RVL), and the corresponding pose of the sensor (represented with a transformation matrix).
Messages are generated at 30Hz.
With the help of the pose of the sensor, the mapping algorithm never fails, and each pixel of all received messages either provides information of a new 3D point or reduces the error of a known 3D point.


We measure the quality of the resulting 3D model by comparing the errors with the reference 3D model.
The reference 3D model is generated by processing all messages in an off-line manner.
The error is measured by the RMSE (root mean square error) of the minimal distance between each point of the comparison 3D model and the reference 3D model.
This is a commonly used method to compare two 3D models.
We employ a tool called CloudCompare ~\cite{cloudcompare} to complete the calculation.

\begin{figure}[!t]
\centering
\includegraphics[width=0.9\columnwidth]{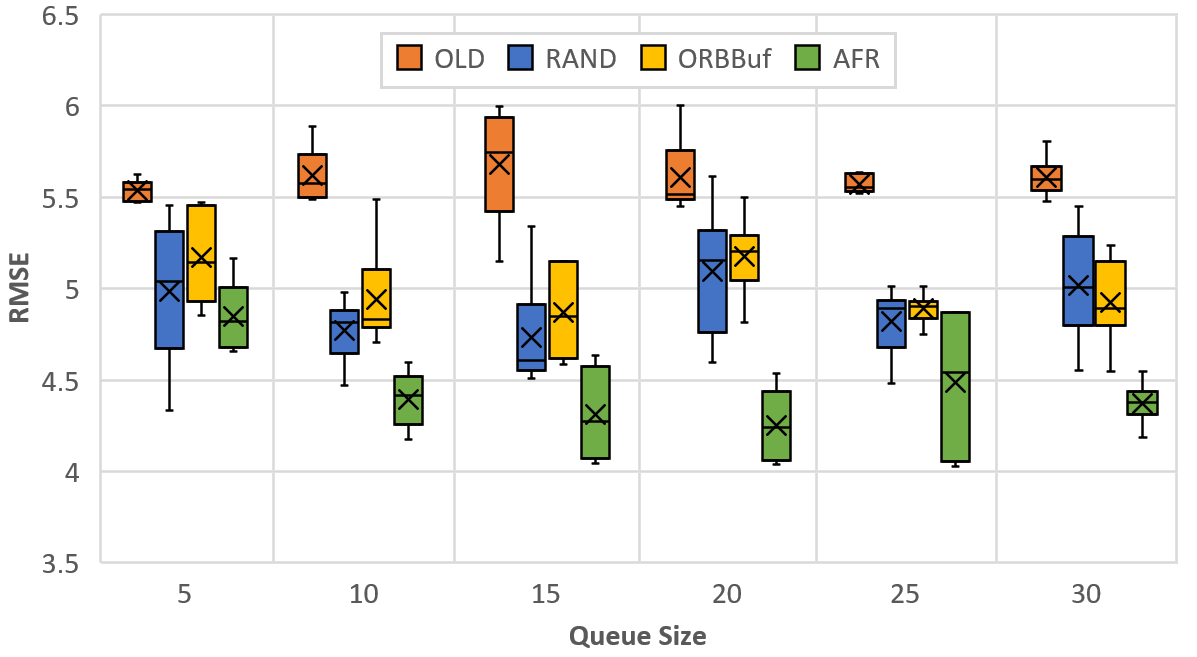}
\caption{The resulting RMSE using different algorithms under different queue sizes.}
\label{fig:rmse}
\end{figure}

The results are organized into Figure~\ref{fig:rmse}.
The horizontal axis is the queue size.
Each box plot indicates the average, the median, the maximum, the minimum, and two quartile values of results of each test case.
Surprisingly, increasing the queue size does not improve the RMSE much, when using the Drop Oldest and the Random Drop algorithm.
This is probably because under this network trace, the queue is full most of the time.
For the ORBBuf algorithm, the situation is not improved, because the ORBBuf algorithm is specialized for tracking algorithms.
For our AFR algorithm, increasing the queue size would reduce the probability of a high dropping rate.
Specifically, when the queue size is 5, our AFR algorithm results in a frame rate of 8 Hz at around the 21st and the 23rd second, which makes the RMSE larger.
On average, our AFR algorithm reduces the average RMSE by about 10\% compared with the Random Drop and the ORBBuf algorithm, and by about 20\% compared with the Drop Oldest algorithm.

\subsection{Remote Tracking Application}
\label{sec:eval-ngt}

\begin{figure}[!t]
\centering
\includegraphics[width=0.9\columnwidth]{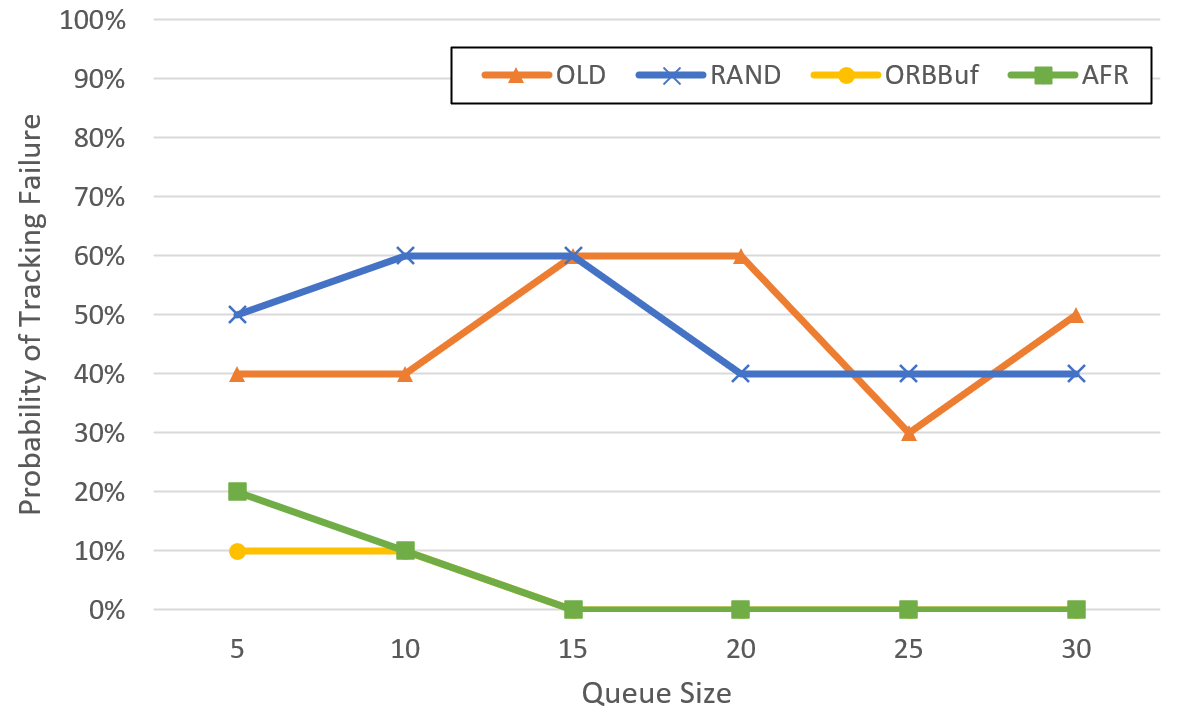}
\caption{Probability of tracking failure using different algorithms under different queue sizes.}
\label{fig:lost}
\end{figure}

In this experiment, we provide the same data sequence, except for the poses of the sensor, to a tracking algorithm~\cite{DBLP:conf/ismar/NewcombeIHMKDKSHF11} under the same network trace condition.
Figure~\ref{fig:lost} shows the probability of tracking failure during our experiments.
The horizontal axis is the queue size.
As we can see, increasing the queue size does not help reducing the probability of tracking failure when using the Drop Oldest and the Random Drop algorithm.
The ORBBuf algorithm is a specified algorithm to meet the needs of tracking algorithms, and the probability of tracking failure is greatly reduced.
Our AFR algorithm can also greatly reduce the probability of tracking failure.
Specifically, when the queue size is larger than 10, no tracking failure occurs.
Overall, our AFR algorithm can reduce the probability of tracking failure from about 40\%-60\% to under 10\%.

Figure~\ref{fig:shot} shows an example of the 3D models resulting from different buffering algorithms.
Figure~\ref{fig:shot_ref} shows the reference reconstructed 3D model.
In Figures~\ref{fig:shot_orig}-\ref{fig:shot_afr}, red points represent points with large error and blue points represent points with small error.
Figure~\ref{fig:shot_orig} shows a result of the Drop Oldest algorithm.
The tracking algorithm failed at around the 23rd second, and its result had already been inaccurate at around the 14th second.
Therefore, its resulting 3D model shows a large error.
Figure~\ref{fig:shot_rand} shows a result of the Random Drop algorithm.
The tracking algorithm failed at around the 7th second.
Therefore, its resulting 3D model loses a lot of information.
Figure~\ref{fig:shot_afr} shows a result of our AFR algorithm.
The tracking algorithm succeeds all the time, and its resulting 3D model is fairly accurate.

\subsection{Time Overhead}
\label{sec:eval-time}

\begin{figure}[!t]
\centering
\includegraphics[width=0.9\columnwidth]{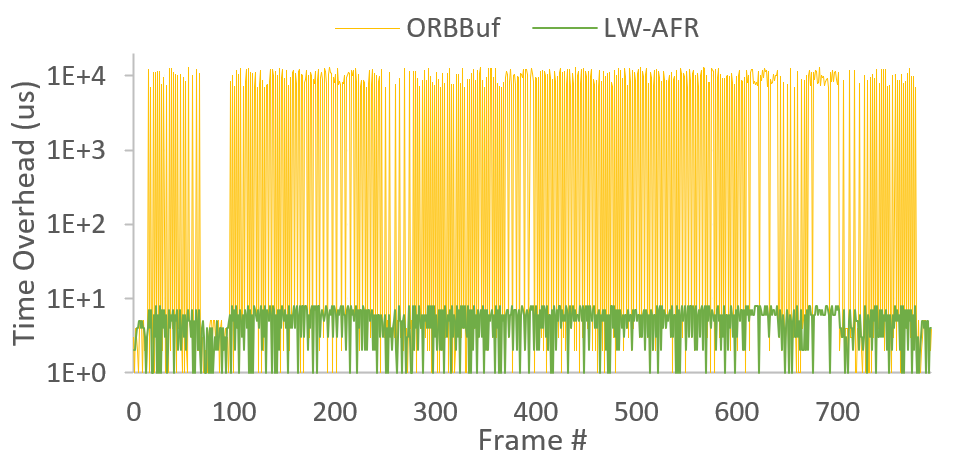}
\caption{Time overhead introduced by different buffering algorithms.
Content-aware algorithms such as ORBBuf introduce more time overhead.
Our AFR algorithm is as efficient as the Drop Oldest and Random algorithms.}
\label{fig:time}
\end{figure}

During the above experiments, we measure the time overhead of different algorithms.
The result is organized in Figure~\ref{fig:time}.
The ORBBuf algorithm introduces a time overhead of about 10 milliseconds when recalculating ORB features is needed.
The network trace used in our experiment is a very challenging one, and the recalculating process is triggered about 400 times within the 794 input frames.
Nonetheless, our AFR algorithm is an efficient algorithm that introduces negligible time overhead (under 10 microseconds), which is of the same order as the Drop Oldest and Random Drop algorithm.
Less time overhead indicates less processor time footprint, so the robot can handle more tasks or consume less energy.

Overall, our experiments have shown that our AFR algorithm can improve the overall performance of cloud robotic systems with negligible time overhead, under challenging network situations where the network bandwidth varies dramatically and data loss is inevitable.

\section{Conclusion}

In this paper, we address the problem caused by data loss during robot-cloud communication.
By modeling the QoI of the data received at the server under the cloud robotic system architecture, we have proposed a novel buffering algorithm that can select the message that can maximize the received QoI.
It is easy to implement and efficient, without the need for feedback from the server, nor heavy and domain-specific content-related calculations.
We have shown that our AFR algorithm performs better than prior algorithms in multiple applications with negligible time overhead.

\bibliographystyle{IEEEtran}
\bibliography{main}

\newpage

\appendices

\section{Proof of Theorems}\label{sec:proof}

By our AFR algorithm, there exists some non-negative integers $x_1$, $x_2$, $r_1$, $r_2$, $n$ and $l$ that satisfy:
\[\begin{split}
    &x_1+x_2+1=L+1 {}\\
    &r_1 = 2^{n+1}, r_2 = 2^n {}\\
    &x_1r_1+x_2r_2+l=T+1 {}\\
    &0<l\leq 2^n
\end{split}\]

Based on the above definition, we get:
\[H_{AFR}(T, L)=x_1V(r_1)+x_2V(r_2)+V(l) \]


\subsection{Lemma 1}

$\forall d_1,d_2,d_3,d_4\geq 1$, if $d_1+d_2=d_3+d_4$ and $|d_1-d_2|<|d_3-d_4|$, then $V(d_1)+V(d_2)\geq V(d_3)+V(d_4)$.

\begin{proof}

Without losing generality, we can assume that $d_3<d_1<d_2<d_4$. Since $V(d)$ is non-convex, by Jensen's Inequality, we have:
\[\begin{split}
    &V(d_1) \geq \frac{d_4-d_1}{d_4-d_3} V(d_3)+\frac{d_1-d_3}{d_4-d_3} V(d_4)  \\
    &V(d_2) \geq \frac{d_4-d_2}{d_4-d_3} V(d_3)+\frac{d_2-d_3}{d_4-d_3} V(d_4)
\end{split}\]

Note that $(d_4-d_1)+(d_4-d_2)=(d_4-d_1)+(d_1-d_3)=d_4-d_3$ and $(d_1-d_3)+(d_2-d_3)=(d_1-d_3)+(d_4-d_1)=d_4-d_3$.
Thus, summing the above inequalities leads to:
\[ V(d_1)+V(d_2) \geq V(d_3)+V(d_4). \]
\end{proof}

\subsection{Theorem 1: No solution is better than that of the $Oracle$'s.}
\label{sec:thm_oracle}

\begin{proof}

For any solution $M$, from our definition of the quantity of information, the distance $d_i$ between adjacent messages satisfies:
\[\begin{split}
    &\sum_{i=1}^{L+1} d_i = T+1 {}\\
    &H_M(T,L)=\sum_{i=1}^{L+1} V(d_i)
\end{split}\]
By Jensen's Inequality, we have
\[
    \frac{\sum_{i=1}^{L+1} V(d_i)}{L+1}\leq V(\frac{\sum_{i=1}^{L+1} d_i}{L+1})=V(\frac{T+1}{L+1})
\]
which shows that $H_S(T,L)\leq H_{Oracle}(T,L)$.

\end{proof}

\subsection{Theorem 2: $2(\sqrt{2}-1)\frac{L}{L+1} < \frac{H_{AFR}(T, L)}{H_{Oracle}(T, L)}\leq 1$}
\label{sec:thm_afr_oracle}

\begin{proof}

1) The left part. Let $\hat{l}=\frac{T+1}{L+1}=\frac{x_1r_1+x_2r_2+l}{x_1+x_2+1}$.
We have $r_1\geq \hat{l} \geq r_2$, and therefore:
\[\begin{split}
    &\frac{H_{AFR}(T, L)}{H_{Oracle}(T, L)} {}\\
    =&\frac{x_1V(r_1)+x_2V(r_2)+V(l)}{(x_1+x_2+1)V(\hat{l})} {}\\
    \geq&\frac{x_1V(\hat{l}) + x_2\frac{r_2}{\hat{l}}V(\hat{l})+0}{(x_1+x_2+1)V(\hat{l})} {}\\
    =&\frac{x_1+x_2\frac{r_2}{\hat{l}}}{x_1+x_2+1} {}\\
    =&\frac{x_1}{L+1}+\frac{L-x_1}{L+x_1+1}
\end{split} \]

By taking the partial derivative of the last term with respect to $x_1$ to be 0, we can get its minimum which leads to:
\[\frac{x_1}{L+1}+\frac{L-x_1}{L+x_1+1} > 2(\sqrt{2}-1)\frac{L}{L+1}\]

2) The right part.
\[\begin{split}
    &H_{Oracle}(T, L){}\\
    =&(L+1)V(\frac{T+1}{L+1}) {}\\
    =&(x_1+x_2+1)V(\frac{x_1r_1+x_2r_2+l}{x_1+x_2+1}) {}\\
    \geq &(x_1+x_2+1)\frac{x_1V(r_1)+x_2V(r_2)+V(l)}{x_1+x_2+1} {}\\
    =&x_1V(r_1)+x_2V(r_2)+V(l) {}\\
    =&H_{AFR}(T, L)
\end{split} \]

\end{proof}

\subsection{Theorem 3: $H_{AFR}(T, L)\geq H_{OLD}(T, L)$.}
\label{sec:thm_afr_old}

\begin{proof}

By Lemma 1 and Jensen Inequality,
\[\begin{split}
    &H_{OLD}(T, L) {}\\
    =&V(x_1(r_1-1)+x_2(r_2-1)+l)+(x_1+x_2)V(1){}\\
    =&V(x_1(r_1-1)+x_2(r_2-1)+l)+V(1){}\\
     &+(x_1+x_2-1)V(1) {}\\
    \leq &V(x_1(r_1-1)+(x_2-1)(r_2-1)+l)+V(r_2){}\\
     &+(x_1+x_2-1)V(1) {}\\
    ... {}\\
    \leq &V(x_1(r_1-1)+l)+x_2V(r_2)+x_1V(1) {}\\
    =&V(x_1(r_1-1)+l)+V(1) {}\\
     &+(x_1-1)V(1)+x_2V(r_2) {}\\
    \leq &V((x_1-1)(r_1-1)+l)+V(r_1) {}\\
     &+(x_1-1)V(1)+x_2V(r_2) {}\\
    ...{}\\
    \leq &V(l)+x_1V(r_1)+x_2V(r_2) {}\\
    =& H_{AFR}(T, L)
\end{split} \]
\end{proof}

\end{document}